\documentclass[11pt]{article}
\usepackage{dialogue2021}
\usepackage{graphicx}
\usepackage{xcolor}
\usepackage{rotating}
\usepackage{ifxetex}
\ifxetex
    \usepackage{fontspec}
    \setromanfont{Times New Roman}
\else
  \usepackage[T1]{fontenc}
  \usepackage[utf8]{inputenc}
  \usepackage{cmap}
  \usepackage{times}
  \usepackage{latexsym}

\fi

\usepackage{amsmath}
\usepackage{amsfonts}
\usepackage{amssymb}

\usepackage[russian,british]{babel}
\usepackage{url}
\usepackage{pgf}

\usepackage{covington} 
\usepackage{etoolbox}

\dialogfinalcopy 

\title{Russian SuperGLUE 1.1: Revising the Lessons\\ not Learned by Russian NLP-models}

\author{Alena Fenogenova$^1$ \\
Tatiana	Shavrina$^{1,2,3}$\\
Alexandr	Kukushkin$^5$\\

  \And
  Maria	Tikhonova$^{1,2}$ \\
  Anton	Emelyanov$^{1,4}$\\
  Valentin	Malykh$^{6,7}$\\
  $^1$SberDevices, Sberbank, Moscow, Russia\\
$^2$National Research University Higher School of Economics, Moscow, Russia\\
$^3$ANO «AI Research Institute», Moscow, Russia\\
$^4$Moscow Institute of Physics and Technology, Moscow, Russia\\
$^5$Alex Kukushkin Lab, Moscow, Russia\\
$^6$Huawei Noah's Ark lab, Moscow, Russia\\
$^7$Kazan Federal University, Kazan, Russia\\[1mm]

   \And
   Vladislav	Mikhailov$^{1,2}$\\
   Denis	Shevelev$^1$\\
   Ekaterina	Artemova$^{2,6}$}

\begin{document}
\maketitle
\bigskip
\bigskip
\bigskip
\bigskip
\bigskip
\begin{abstract}
  
  In the last year, new neural architectures and multilingual pre-trained models have been released for Russian, which led to performance evaluation problems across a range of language understanding tasks. 

This paper presents Russian SuperGLUE 1.1, an updated benchmark styled after GLUE for Russian NLP models. The new version includes a number of technical, user experience and methodological improvements, including fixes of the benchmark vulnerabilities unresolved in the previous version: novel and improved tests for understanding the meaning of a word in context (RUSSE) along with reading comprehension and common sense reasoning  (DaNetQA, RuCoS, MuSeRC). Together with the release of the updated datasets, we improve the benchmark toolkit based on \texttt{jiant} framework for consistent training and evaluation of NLP-models of various architectures which now supports the most recent models for Russian.
Finally, we provide the integration of  Russian SuperGLUE with a framework for industrial evaluation of the open-source models, MOROCCO (MOdel ResOurCe COmparison), in which the models are evaluated according to the weighted average metric over all tasks, the inference speed, and the occupied amount of RAM.
Russian SuperGLUE is publicly available at \url{https://russiansuperglue.com/}.
  
  \textbf{DOI:} 10.28995/2075-7182-2021-20-XX-XX
\end{abstract}

\selectlanguage{russian}
\begin{center}
  \russiantitle{Russian SuperGLUE 1.1: пересматривая невыученные уроки\\русскоязычных NLP-моделей}

\medskip \setlength\tabcolsep{0.82cm}
  \begin{tabular}{ccc}
    \textbf{Алена Феногенова$^1$} & \textbf{Мария Тихонова$^{1,2}$} & \textbf{Владислав Михайлов$^{1,2}$}\\
      Татьяна Шаврина$^{1,2,3}$ & Антон Емельянов$^{1,4}$ & Денис Шевелев$^1$\\
      Александр Кукушкин$^5$ & Валентин Малых$^{6,7}$ & Екатерина Артемова$^{2,6}$\\
  \end{tabular}
  \medskip

$^1$SberDevices, Сбербанк, Москва, Россия\\[1mm]
$^2$НИУ «Высшая школа экономики», Москва, Россия\\[1mm]
$^3$АНО «Институт Искусственного Интеллекта», Москва, Россия\\[1mm]
$^4$Московский физико-технический институт, Москва, Россия\\[1mm]
$^5$Лаборатория Александра Кукушкина, Москва, Россия\\[1mm]
$^6$Huawei Noah's Ark lab, Москва, Россия\\[1mm]
$^7$Казанский (Приволжский) федеральный университет, Казань, Россия\\[1mm]
\end{center}
\bigskip
\bigskip
\bigskip
\begin{abstract}
  В прошлом году на русскоязычном материале были обучены новые нейронные архитектуры, в том числе мультиязычные NLP-модели, что привело к новым вызовам в оценке качества решений задач понимания естественного языка.

В этой статье представлен Russian SuperGLUE 1.1, бенчмарк на основе GLUE для оценки языковых моделей для русского языка. Новая версия включает в себя 
ряд технических обновлений, улучшение пользовательского опыта и устранение методологических уязвимостей версии 1.0., в том числе создание новых тестовых сетов и улучшение датасетов на понимание смысла слова в контексте (RUSSE), машинное чтение и здравый смысл (DaNetQA, RuCoS, MuSeRC). Кроме того, представлены технические обновления бенчмарка на основе фреймворка \texttt{jiant} для консистентного обучения и оценки NLP-моделей различных архитектур, включая самые последние модели для русского языка.
Помимо обновления основного бенчмарка, мы представляем интеграцию бенчмарка Russian SuperGLUE с фреймворком для промышленной оценки моделей с открытым исходным кодом --  MOROCCO (MOdel ResOurCe COmparison), в котором модели оцениваются по средневзвешенной метрике всех заданий, скорости быстродействия и занимаемого объема оперативной памяти. Материалы Russian SuperGLUE доступны по адресу \url{https://russiansuperglue.com/}.
  
  \textbf{Ключевые слова:} оценка моделей, понимание естественного языка, бенчмарки, NLP-модели
\end{abstract}
\selectlanguage{british}

\section{Introduction}

In the last years, new architectures and methods for model pre-training and transfer learning have driven striking performance improvements across a range of language understanding tasks. Complex benchmark approaches are being developed for testing general intellectual ``abilities'' of NLP models on a wide range of natural language understanding (NLU) tasks. The tasks range from identifying causal relations in texts (NLI) to common sense, world knowledge, and logic. The central benchmarks in the field are GLUE [1] and SuperGLUE [2] projects for English, they include versatile tasks and allow competitive evaluation of the models on a public leaderboard. Recently, analogous general language understanding evaluation benchmarks have been developed for Chinese [3], French [4], Polish [5] and Russian [6]. RussianSuperGLUE provides nine novel Russian NLU tasks, a public leaderboard, count-based and transformer-based baselines, and human solver evaluation.

This work presents Russian SuperGLUE 1.1, a new release of the benchmark that provides multiple updates and improvements of the previous version. 
First, we updated the following datasets: 1) RUSSE: expansion of the dataset and construction of a novel test set; 2) DaNetQA: increasing the size of the dataset and creation of a new test set; 3) RuCoS: doubling the size of the validation and test sets, cleaning typos and inaccuracies; 4) MuSeRC: the expansion of the dataset, cleaning typos and inaccuracies.
Second, we provide an improved Russian SuperGLUE toolkit based on \texttt{jiant} framework [7] for consistent training and evaluation of NLP models for Russian, which now supports the novel transformer-based models such as RuGPT\footnote{https://github.com/sberbank-ai/ru-gpts/tree/master}.
Furthermore, we introduce an enhanced web interface of the benchmark that includes bug fixes and new features: the model evaluation by individual task (one can get the score for a specific task), a better notification procedure, and a new leaderboard based upon the model performance evaluation. Finally, Russian SuperGLUE has been integrated with MOROCCO, a framework for industrial evaluation of model performance. Models submitted to the leaderboard can be additionally estimated by inference speed and memory footprint.


The remainder is organized as follows. Section 2 briefly describes the benchmark tasks. Section 3 outlines the new release, namely the dataset updates and improvements of the leaderboard interface. Section 4 provides the description of MOROCCO framework and the performance evaluation metrics. We compare a number of novel models for Russian with English ones in Section 5 and conclude in Section 6.


\section{Previous Work}
\label{sec:background}
Russian-based NLP-systems have a long history of benchmarking within various tasks. Starting with ROMIP Seminar in 2003\footnote{\url{http://romip.ru/ru/2003/index.html}}, then Dialog Evaluation tracks starting from 2008\footnote{\url{http://www.dialog-21.ru/evaluation/}} have continued the prolific tradition of yearly system evaluation on the most technically relevant problems, including morphological and syntactic parsing, text classification, spell check, named entity recognition, and many more. RUSSE'2018\footnote{\url{https://russe.nlpub.org/2018/wsi/}}, word sense induction and disambiguation for the Russian shared task, is definitely worth mentioning as well. Last but not least, SberSQuAD [8] QA-system leaderboard completes a series of traditional single-task benchmarks. 

Russian SuperGLUE benchmark first introduced a multi-task benchmark for Russian, providing a stable updated leaderboard with all the systems ranged by their average performance on 9 complex tasks. 
 
\subsection{Russian SuperGLUE Tasks}
We continue our work on Russian SuperGLUE\footnote{ \url{https://russiansuperglue.com/}} [6] which follows the general language understanding evaluation methodology. Similarly to the English prototype, Russian benchmark includes a set of NLU tasks and a publicly available leaderboard. Namely, the benchmark comprises 9 tasks divided into 5 groups:
\begin{itemize}
    \item \textbf{Textual Entailment \& NLI}: TERRa, RCB, LiDiRus;
    \item \textbf{Common Sense}: RUSSE, PARus;
    \item \textbf{World  Knowledge}: DaNetQA [9];
    \item \textbf{Machine Reading}: MuSeRC, RuCoS [10];
    \item \textbf{Reasoning}: RWSD.
\end{itemize}

\paragraph{Task Description} We outline the information on the tasks by their type, metrics and partition sizes in  Table~\ref{table:tasks}.

\begin{table*}[]
\centering
\begin{tabular}{|c|c|c|c|c|c|}
\hline
\textbf{Task} & \textbf{Task Type} & \textbf{Task Metric} & \textbf{Train} & \textbf{Val} & \textbf{Test} \\ \hline
TERRa         & NLI                & Accuracy             & 2616           & 307          & 3198          \\ \hline
RCB           & NLI                & Avg. F1 / Accuracy   & 438            & 220          & 438           \\\hline
LiDiRus       & NLI \& diagnostics & MCC                  & 0              & 0            & 1104          \\ \hline
RUSSE         & Common Sense       & Accuracy             & 19845          & 8508         & 18892         \\ \hline
PARus         & Common Sense       & Accuracy             & 400            & 100          & 500           \\ \hline
DaNetQA       & World Knowledge    & Accuracy             & 1749           & 821          & 805           \\ \hline
MuSeRC        & Machine Reading    & F1 / EM              & 500            & 100          & 322           \\ \hline
RuCoS         & Machine Reading    & F1 / EM              & 72193          & 7 577        & 7257          \\ \hline
RWSD          & Reasoning          & Accuracy             & 606            & 204          & 154           \\ \hline
\end{tabular}
\caption{Russian SuperGLUE task description. Train/Val/Test include number of samples for each set; MCC stands for Matthews Correlation Coefficient; EM - Exact Match.} \label{table:tasks}
\end{table*}

\textbf{TERRA} Textual Entailment Recognition for Russian is aimed at capturing textual entailment in a binary classification form. Given two text fragments (premise and hypothesis), the task is to determine whether the meaning of the hypothesis is entailed from the premise. The dataset was sampled from the Taiga corpus [11].

\textbf{RCB} The Russian Commitment Bank is a 3-way classification task aimed at recognizing textual entailment (NLI).  In contrast to TERRa, the premise in RCB may represent a textual segment rather than a single sentence. The corpus was filtered from Taiga with a number of pre-defined rules and labeled by crowd workers.  

\textbf{LiDiRus} is a diagnostic set that tests models for a rich set of 33 linguistic features, commonsense, and world knowledge. The dataset was constructed as a translation from GLUE diagnostics with the preservation of all features. Thus, it provides an opportunity for evaluation of linguistic and semantic properties of language models in the setting of NLI task and for drawing comparisons between the languages.

\textbf{RUSSE} is a binary classification task that involves word sense disambiguation. Given a pair of sentences containing the same ambiguous word, the goal of the model is to recognize if the word is used in the same meaning. The dataset was constructed from RUSSE [12].

\textbf{PARus} is a binary classification task aimed to identify the most plausible alternative out of two for a given premise. It is a manually verified translation of the COPA dataset from SuperGLUE. 

\textbf{DaNetQA} is a Russian QA dataset for yes/no questions which follows the BoolQ design [21]. Each sample consists of a Wikipedia paragraph and a human-generated question related to the paragraph. The task is to come up with a binary answer (yes or no) for the given question.

\textbf{MuSeRC} is a machine reading comprehension (MRC) task. Each sample consists of a text paragraph, multi-hop questions based on the paragraph, and possible answers for each question. The goal is to choose all correct answers for each question. The dataset was collected from publicly available sources across multiple domains (elementary school texts, news, summary of series, fiction stories, and fairy tales), and further annotated by crowd-workers.

\textbf{RuCoS} is an MRC task that involves commonsense reasoning and world knowledge. The dataset is a counterpart of ReCoRD [22] for English. Each example consists of a text paragraph, a query with a missing named entity, and a set of candidates for the answer. The task is to select one of the candidates that best fits the gap.

\textbf{RWSD} Russian Winograd Schema task is devoted to coreference resolution in a binary classification form. The corpus was created as a manually validated translation of the Winograd Schema Challenge\footnote{\url{https://cs.nyu.edu/faculty/davise/papers/WinogradSchemas/WS.html}}.







\section{New Release Features}
Version 1.1 includes important methodological updates to the datasets, as well as the expansion of a number of "out-of-the-box" supported model architectures in the software, which is attached for the convenience of developers and a unified testing environment for all systems. Also, in addition to the main leaderboard, the model evaluation process was significantly supplemented by industrial metrics, which will be described below.

\subsection{Improving the Tasks}
The first version of the datasets has a number of drawbacks that revealed themselves after the initial release. We collected the feedback and fixed the shortcomings of the previous version of the benchmark. Among the main reasons for weaknesses are: 1) data leakage, 2) class distributions in test sets, 3) smaller size of the MRC datasets and the number of typos and inconsistencies in them. The latest problem is inevitable, RuCoS and MuSeRC are the most resources and time-consuming for collection and verification datasets - still their sizes were smaller than in their English analogs (ReCoRD and MultiRC respectively).
As a result, four datasets were improved, for two of them (DaNetQA and RUSSE) completely new test sets were created. The results of the leaderboard were rescored - all baselines were measured again on the new datasets. Additionally, we asked the participants to resubmit on the new data. Thus, now only the latest version of the datasets and leaderboard are supported. 
In this section, we describe the procedure of the dataset improvements.

\subsubsection{RUSSE}

The update of RUSSE was motivated by extremely high scores of the language models which significantly outperformed the human benchmark.

For example, mBART [28] achieved almost 99\% accuracy, while human performance was at the level of 75\%. We believe that the conversion of the publicly available RUSSE test set to that of Russian SuperGLUE has possibly led to data leakage. 

To this end, we constructed a completely new test set for the task in order to eliminate the potential leakage. First, we filtered the anchor words, discarded the most outdated and rare ones, and enriched the dataset with novel samples. Second, we collected sentences dating from the 2020 year using publicly available news sources such as Wikinews\footnote{https://www.wikinews.org/} and Lenta.ru\footnote{https://lenta.ru/}. Finally, we manually validated the meanings of the anchor words in the resulted sentences and annotated the answers. The total size of the novel test set is 18 892 examples.

For the obtained test set we re-scored the human benchmark using the same annotation procedure in Yandex.Toloka task as described in [6] but on the new subset of the data. 
The human performance achieved 80.5\% accuracy, while the best model performance on the leaderboard~\ref{tab:leaderboard} at present is 72.9\% (RuBERT conversational). 

\subsubsection{DaNetQA}
Originally, DaNetQA had a limited number of examples: 392, 295, 295 (train/val/test). We extended the dataset following the methodology described in [9], and converted a subset of MuSeRC into the yes/no QA setting, labeled by crowd-workers afterward.
The new task contains 1750, 821, and 805 examples (train/val/test). In addition, we manually checked validation and test sets and balanced both sets by target class, as opposed to the previous version where the class distribution was 80/20\% and changed the answer distribution by balancing the sets. The current class balance is 50/50\% in contrary to the originally imbalanced data with 80\% yes answers.

Since the test set has been changed completely we re-scored the performance of human solvers and models. While human performance gets 91 of accuracy score for the updated dataset, the language models (see table~\ref{tab:leaderboard}) are not greater than 65,7\% (not 80\% as it was before).

\subsubsection{RuCoS}
The new version of RuCoS involves the following updates. We doubled the size of the validation (7527 examples) and test (7257 examples) sets as described in [10]. We manually verified the crowd-worker annotations and corrected typos and annotation inconsistencies. Since the human performance was assessed on a subset of the test set, the results remain the same. The best-performing model is now RuGPT3-XL over a few-shot technique.

\subsubsection{MuSeRC}
As opposed to the English analogue MultiRC, MuSeRC was relatively small in size which we aimed to improve. However, MuSeRC consists only of multi-hop questions which makes the tasks more difficult in contrast to MultiRC which also includes one-hop questions. We extended the train set with more than 300 new samples containing novel multi-hop questions. As a result, the size of MuSeRC became comparable with MultiRC as the number of multi-hop questions is 5,228 and 5,825 respectively. 
Thus, in the new Russian SuperGLUE release we: expanded the train set; cleaned typos, grammar mistakes, and text inaccuracies in all the samples.

\subsection{Infrastructure Advances}
The interaction between the leaderboard participants and the benchmark interface is crucially important as the entry usage threshold directly affects the user experience and submission quantity. To this end, the new release contains bug fixes and presents new features of the leaderboard interface and infrastructure.

\begin{figure}[hbt!]
    \centering
    \includegraphics[width=\linewidth]{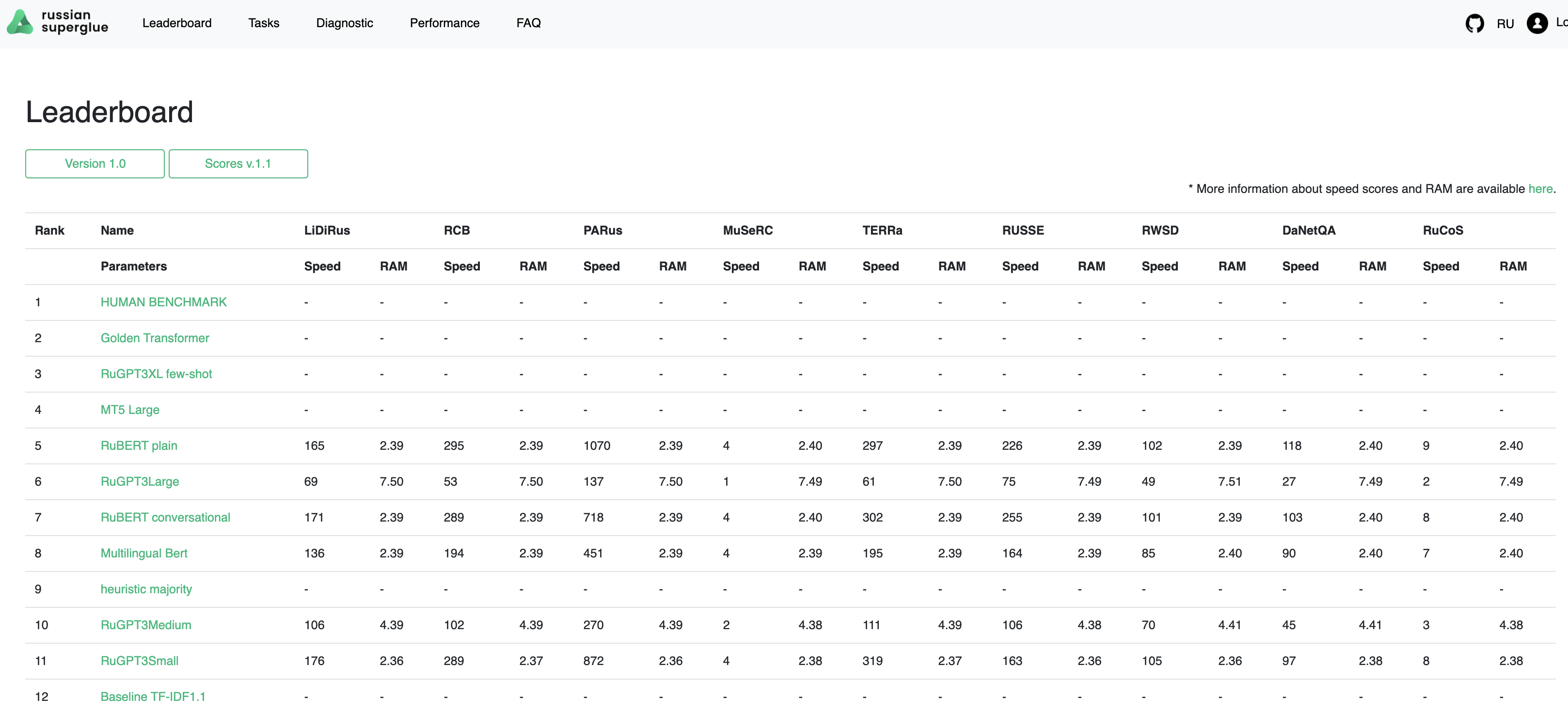}
    \caption{Performance evaluation leaderboard in RussianSuperGLUE benchmark.}
    \label{fig:performance_leaderboard}
\end{figure}

First, we improved the reliability of the model evaluation system on the website. We made it more strict and fixed some minor bugs. Furthermore, we enhanced the web interface and added new features:
1)	The user can download their previous submissions and edit them;
2) 	The user can evaluate their model and upload submission both on a single task and the full set of tasks;
3) 	The user receives two email notifications after they make their submission public. The first one confirms the submission verification step, and the second one informs whether the submission was published on the leaderboard or rejected (and why);
4) The user guide is updated to provide a better leaderboard usage experience;
5) The user now can evaluate their model by industrial performance metrics, namely inference speed and memory footprint which we describe in Section 4. The performance leaderboard is developed as well (see Figure~\ref{fig:performance_leaderboard}).

Besides, Russian SuperGLUE 1.1 involves minor bug fixes along with the support of the novel models for Russian: RuGPT3 models\footnote{
RuGPT3-Small: \url{https://huggingface.co/sberbank-ai/rugpt3small_based_on_gpt2}, \\
RuGPT3-Medium: \url{https://huggingface.co/sberbank-ai/rugpt3medium_based_on_gpt2},\\
RuGPT3-Large: \url{https://huggingface.co/sberbank-ai/rugpt3large_based_on_gpt2}, \\
RuGPT3-XL: \url{https://huggingface.co/sberbank-ai/rugpt3xl}
}
included in the list of models by HuggingFace library\footnote{\url{https://github.com/huggingface/transformers}}.

\section{Computational Efficiency Evaluation}
A number of closely related benchmarks consider only the task-specific performance of the models, leaving the computational efficiency unattended.
To this end, Russian SuperGLUE is now integrated with MOROCCO [31], a framework for industrial evaluation of model performance by the following metrics: \textit{memory footprint} and \textit{inference speed} (see Section \ref{sec:metrics}). The user can evaluate their model by submitting a Docker container which is expected to read the text from the standard input and channel the predictions to standard output. The container is run in a fixed isolated environment with limited running time, RAM, and CPU/GPU resources. We use Yandex.Cloud\footnote{\url{https://cloud.yandex.com/}} platform where the following hardware is provided: 1 $\times$ Intel Broadwell CPU, 1 $\times$ NVIDIA Tesla V100 GPU. The Docker container OS is Ubuntu 20.04. The solution is run over several iterations to eliminate the  dispersion, with the median values further computed. Along with the metrics, we also compute the task-specific metric based upon the Docker output to further aggregate the results into a final score for the submission.

\subsection{Industrial Metrics}
\label{sec:metrics}

\textit{Memory footprint}, or GPU RAM usage $M$ is measured by running a Docker container with a single record as input and measuring the maximum $M$. We repeat the procedure 5 times and take the median value.

\vspace{0.5em}\noindent \textit{Inference speed}, or \emph{throughput} $Tp$ is computed by running a Docker container with $N$ records as input and optional batch size to measure $T_N$. Besides, we estimate the model initialization time $T_{\text{init}}$ by running the container with a single record as input. The resulting \emph{throughput} is computed as follows\footnote{During the evaluation process, the $N=2000$ is used and the batch size is 32. We repeat the procedure 5 times and compute the median value.}: $$Tp = \frac{N}{T_{N} - T_{\text{init}}}.$$

\begin{figure}[hbt!]
    \centering
    \includegraphics[width=0.7\textwidth]{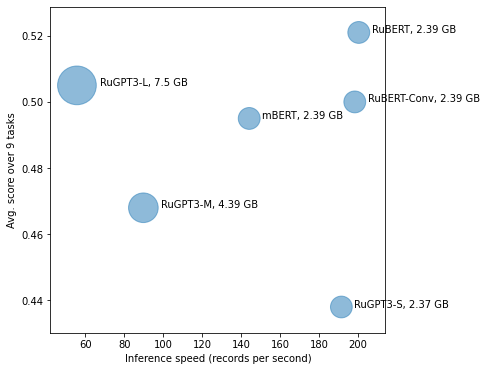}
    \caption{Models comparison on Russian SuperGLUE benchmark.}
    \label{fig:russianglue}
\end{figure}

 We propose to use these three characteristics, namely $Q$, $Tp$, and $M$, in the following way: we comprise a 2-dimensional plot with the horizontal axis being a quality for a downstream task $Q$ (this metric is specific to the task) and vertical axis being a throughput $Tp$ for the model. To visualize memory footprint $M$, we propose to use circles of different sizes instead of a mere point on the plot. The scores that involve industrial performance metrics for the models from Russian SuperGLUE leaderboard are presented in Figure~\ref{fig:russianglue}.

MOROCCO allows computing the performance by running the containers in other technical environments that best fit the user needs. Figure 2 presents an unexpected result: the dilemma of choosing between speed and performance for the Russian language is not at all a dilemma - the highest quality models are also the fastest (certainly, except for the simplest solutions and baselines like TF-IDF).

\section{Results and Discussion}
The current level of systems participating in the above-described language tests and competitions has certainly grown over the past decades: although the most common text-based tests are less than a dozen years old, we can trace the development trends and system evaluations based on simpler, unchanged technical criteria for their growth. 
The benchmark approach to the assessment of intelligent systems is currently dominant, allowing to combine the assessment of various intellectual abilities under the cumulative assessment of general intelligence. Intellectual tests, expressed through texts, constitute the main productive method of such an assessment, making it possible to formulate a variety of types of tasks and compare the level of systems with human intelligence, including the formation of sets of examples of tasks, to successfully solve skills or abilities that are not lower than human, but which do not have clear definitions within the framework of neuroscience: common sense, goal-setting, cause-and-effect relationships, knowledge about the world. The current version of the leaderboard version 1.1 is shown in Figure 3.

The existing problems of benchmark approaches, however, are subject to close research by the community. The discussion is provoked by a significant difference in the level of metrics observed in Russian SuperGLUE and in the English-language leaderboard. The best result so far is 67.9\% overall for Russian (while the human level is 81.1\%), while for English the best model performance is 90.4\%, beating the human score of 89.8\%.

A separate subject of discussion is the issue of limitations of the presented leaderboard and its methodological analogues in other languages. While the tasks set in the benchmark themselves are designed to test the human intellectual abilities or their imitation, we see that in some cases (for example, the English SuperGLUE), a result higher than human has already been achieved in a completely mechanical approach, using transformer models pretrained on large corpora and fine-tuned on accumulated task-specific data. At the time of this writing, the best result among Russian-based models was raised by 15\% using simplistic ML-hackathon methods - model ensembling, automatic translation and training a meta-classifier for weighting the models in various tasks.

The growing popularity of multilingual benchmarks is promising: the extension of the testing methodology has led to a comprehensive multilingual assessment - the XTREME [29] and XGLUE [30] projects have combined the available materials for testing systems to their ability to reproduce human intellectual abilities on 40 and 19 typologically diverse languages accordingly. The Russian language is included as a part of the test material in both of these benchmarks, which is a promising prospect for the further integration of project data into multilingual X-benchmarks with both testing and training data provided. We anticipate that the speed of benchmark hacking will become lower than the speed of creating weighted, complex benchmarks if they strictly evaluate the modelling of the language and separate the language itself from all distinct abilities expressed with the language.

\begin{figure}[h!]
    \centering
    \includegraphics[width=\linewidth]{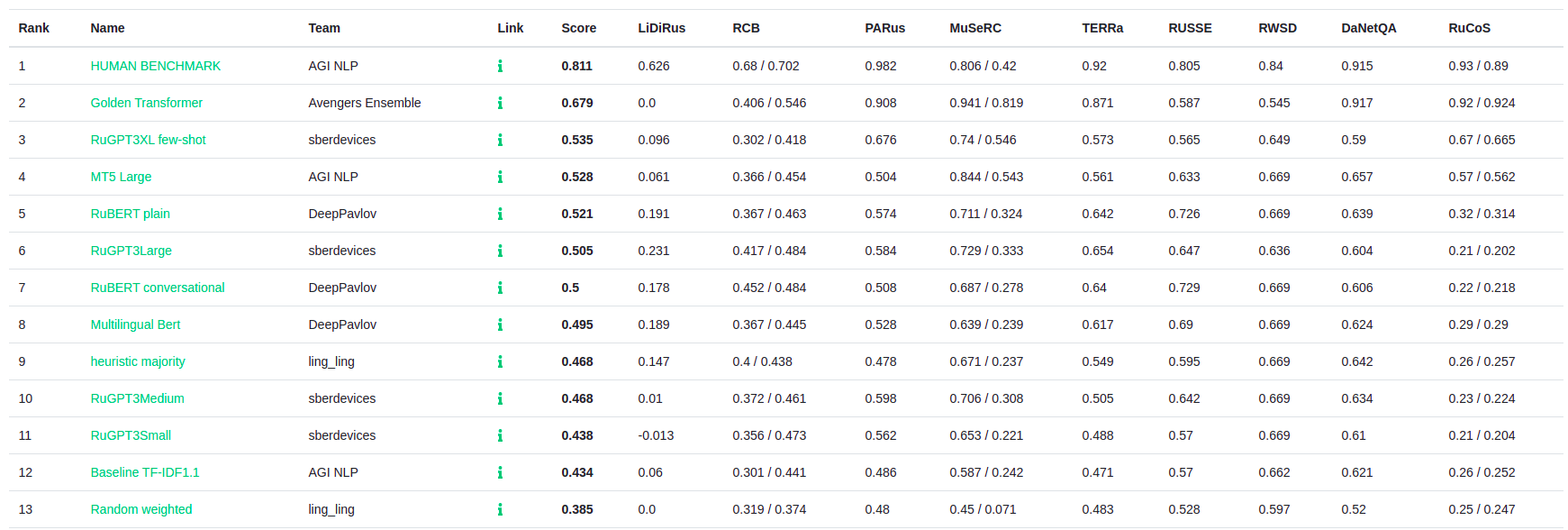}
    \caption{Models comparison on RussianSuperGLUE benchmark.}
    \label{fig:russianglue_main}
\end{figure}

\section{Conclusion}
We present Russian SuperGLUE v1.1, an updated benchmark for evaluating general-purpose language understanding systems for Russian. 
As part of the development of the project, the tasks were updated:
\begin{itemize}
\item RUSSE (understanding word meaning in context) - a new test set has been compiled, the possibility of a dataset leak is excluded;
\item DaNetQA (yes / no questions based on commonsense reasoning and machine reading) - a new test set was compiled, the composition of classes in the dev and test set was balanced;
\item RuCoS (machine-reading and commonsense reasoning) - task dataset expanded and manually corrected, the composition of the classes in the dev and test set has been balanced;
\item MuSeRC (machine-reading and information retrieval) - task dataset expanded and manually corrected.
\end{itemize}
A prominent direction for future work is to expand the existing tasks, provide the support for upcoming models for Russian, and improve the user experience of the MOROCCO framework, specifically by supporting models not released as a part of HuggingFace library such as ELMo\footnote{\url{http://docs.deeppavlov.ai/en/master/features/pretrained_vectors.html#elmo}}. Overall, we believe that Russian SuperGLUE provides the research community with a challenging frontier and further natural language understanding progress for Russian.

\color{blue}\section*{References}
\color{black}

\hspace{0.27cm} [1] Wang, Alex, et al. ``GLUE: A multi-task benchmark and analysis platform for natural language understanding.'' arXiv preprint arXiv:1804.07461 (2018).\\

[2] Wang, Alex, et al. ``Superglue: A stickier benchmark for general-purpose language understanding systems.'' arXiv preprint arXiv:1905.00537 (2019).\\

[3] Xu L. et al. Clue: A chinese language understanding evaluation benchmark //arXiv preprint arXiv:2004.05986. – 2020.\\

[4] Le H. et al. Flaubert: Unsupervised language model pre-training for french //arXiv preprint arXiv:1912.05372. – 2019.\\

[5] Rybak P. et al. KLEJ: comprehensive benchmark for polish language understanding //arXiv preprint arXiv:2005.00630. – 2020. \\

[6] Shavrina, Tatiana, et al. ``RussianSuperGLUE: A Russian Language Understanding Evaluation Benchmark.'' arXiv preprint arXiv:2010.15925 (2020).\\

[7] Pruksachatkun, Yada, et al. "jiant: A software toolkit for research on general-purpose text understanding models." arXiv preprint arXiv:2003.02249 (2020).\\

[8] Efimov P. et al. SberQuAD–Russian reading comprehension dataset: Description and analysis //International Conference of the Cross-Language Evaluation Forum for European Languages. – Springer, Cham, 2020. – pp. 3-15.\\

[9] Glushkova, Taisia, et al. ``DaNetQA: a yes/no Question Answering Dataset for the Russian Language.'' arXiv preprint arXiv:2010.02605 (2020).\\

[10] Fenogenova, Alena, Vladislav Mikhailov, and Denis Shevelev. ``Read and Reason with MuSeRC and RuCoS: Datasets for Machine Reading Comprehension for Russian.'' Proceedings of the 28th International Conference on Computational Linguistics. 2020.\\

[11] Shavrina, Tatiana, and Olga Shapovalova. ``To the methodology of corpus construction for machine learning: “Taiga” syntax tree corpus and parser.'' Proceedings of ``CORPORA-2017'' International Conference. 2017.\\

[12] Panchenko, Alexander, et al. ``RUSSE'2018: a shared task on word sense induction for the Russian language.'' arXiv preprint arXiv:1803.05795 (2018).\\

[13] Sharoff, Serge, and Joakim Nivre. ``The proper place of men and machines in language technology." Processing Russian without any Linguistic Knowledge. Computational Linguistics and Intelligent Technologies 10.17 (2011): 657-670.\\

[14] Conneau, Alexis, and Douwe Kiela. ``Senteval: An evaluation toolkit for universal sentence representations.'' arXiv preprint arXiv:1803.05449 (2018).\\

[15] McCann, Bryan, et al. ``The natural language decathlon: Multitask learning as question answering.'' arXiv preprint arXiv:1806.08730 (2018).\\

[16] Eichler, Max, Gözde Gül Şahin, and Iryna Gurevych. ``LINSPECTOR WEB: A multilingual probing suite for word representations.'' arXiv preprint arXiv:1907.11438 (2019).\\

[17] Şahin, Gözde Gül, et al. ``Linspector: Multilingual probing tasks for word representations.'' Computational Linguistics 46.2 (2020): 335-385.\\

[18] Wolf, Thomas, et al. ``HuggingFace's Transformers: State-of-the-art natural language processing.'' arXiv preprint arXiv:1910.03771 (2019).\\

[19] Gauen, Kent, et al. ``Low-power image recognition challenge.'' 2017 22nd Asia and South Pacific Design Automation Conference (ASP-DAC). IEEE, 2017.\\

[20] Panchenko, Alexander, et al. ``Russe: The first workshop on russian semantic similarity.'' arXiv preprint arXiv:1803.05820 (2018).\\

[21] Clark, Christopher, et al. ``BoolQ: Exploring the surprising difficulty of natural yes/no questions.'' arXiv preprint arXiv:1905.10044 (2019).\\

[22] Zhang, Sheng, et al. ``Record: Bridging the gap between human and machine commonsense reading comprehension.'' arXiv preprint arXiv:1810.12885 (2018).\\

[23] Dagan, Ido, Oren Glickman, and Bernardo Magnini. ``The pascal recognising textual entailment challenge.'' Machine Learning Challenges Workshop. Springer, Berlin, Heidelberg, 2005.\\

[24] Haim, R. Bar, et al. ``The second pascal recognising textual entailment challenge.'' Proceedings of the Second PASCAL Challenges Workshop on Recognising Textual Entailment. 2006.\\

[25] Giampiccolo, Danilo, et al. ``The third pascal recognizing textual entailment challenge.'' Proceedings of the ACL-PASCAL workshop on textual entailment and paraphrasing. 2007.\\

[26]  Bentivogli, Luisa, et al. ``The Fifth PASCAL Recognizing Textual Entailment Challenge.'' TAC. 2009.\\

[27] Min, Sewon, et al. ``NeurIPS 2020 EfficientQA Competition: Systems, Analyses and Lessons Learned.'' arXiv preprint arXiv:2101.00133 (2021). \\

[28] Liu, Yinhan, et al. "Multilingual denoising pre-training for neural machine translation." Transactions of the Association for Computational Linguistics 8 (2020): 726-742. \\

[29] Hu J. et al. XTREME: A Massively Multilingual Multi-task Benchmark for Evaluating Cross-lingual Generalisation //International Conference on Machine Learning. – PMLR, 2020. – pp. 4411-4421. \\

[30] Liang Y. et al. Xglue: A new benchmark dataset for cross-lingual pre-training, understanding and generation //arXiv preprint arXiv:2004.01401. – 2020. \\

[31] V.Malykh, A.Kukushkin, E.Artemova, V.Mikhailov, M.Tikhonova, T.Shavrina. MOROCCO: Model Resource Comparison Framework. //arXiv preprint arXiv:2104.14314. – 2021.



\section{Appendix}
\subsection{Appendix 1. Russian SuperGLUE Leaderboard}
\begin{sidewaystable}
\centering
\caption{Russian SuperGLUE v 1.1 Leaderboard}\label{tab:leaderboard}
\begin{tabular}{|l|l|l|l|l|l|l|l|l|l|l|l|}
\hline
Rank & Name                                                         & Score & LiDiRus & RCB           & PARus & MuSeRC        & TERRa & RUSSE & RWSD  & DaNetQA & RuCoS        \\ \hline
1    & \begin{tabular}[c]{@{}l@{}}HUMAN \\ BENCHMARK\end{tabular}   & 0.811 & 0.626   & 0.68 / 0.702  & 0.982 & 0.806 / 0.42  & 0.92  & 0.805 & 0.84  & 0.915   & 0.93 / 0.89  \\ \hline
2    & \begin{tabular}[c]{@{}l@{}}RuGPT3XL\\ few-shot\end{tabular}  & 0.535 & 0.096   & 0.302 / 0.418 & 0.676 & 0.74 / 0.546  & 0.573 & 0.565 & 0.649 & 0.59    & 0.67 / 0.665 \\ \hline
3    & \begin{tabular}[c]{@{}l@{}}MT5\\ Large\end{tabular}          & 0.528 & 0.061   & 0.366 / 0.454 & 0.504 & 0.844 / 0.543 & 0.561 & 0.633 & 0.669 & 0.657   & 0.57 / 0.562 \\ \hline
4    & \begin{tabular}[c]{@{}l@{}}RuBERT\\ plain\end{tabular}       & 0.521 & 0.191   & 0.367 / 0.463 & 0.574 & 0.711 / 0.324 & 0.642 & 0.726 & 0.669 & 0.639   & 0.32 / 0.314 \\ \hline
5    & RuGPT3Large                                                  & 0.505 & 0.231   & 0.417 / 0.484 & 0.584 & 0.729 / 0.333 & 0.654 & 0.647 & 0.636 & 0.604   & 0.21 / 0.202 \\ \hline
6    & \begin{tabular}[c]{@{}l@{}}RuBERT\\ conv\end{tabular}        & 0.5   & 0.178   & 0.452 / 0.484 & 0.508 & 0.687 / 0.278 & 0.64  & 0.729 & 0.669 & 0.606   & 0.22 / 0.218 \\ \hline
7    & mBert                                                        & 0.495 & 0.189   & 0.367 / 0.445 & 0.528 & 0.639 / 0.239 & 0.617 & 0.69  & 0.669 & 0.624   & 0.29 / 0.29  \\ \hline
8    & \begin{tabular}[c]{@{}l@{}}heuristic\\ majority\end{tabular} & 0.468 & 0.147   & 0.4 / 0.438   & 0.478 & 0.671 / 0.237 & 0.549 & 0.595 & 0.669 & 0.642   & 0.26 / 0.257 \\ \hline
9    & \begin{tabular}[c]{@{}l@{}}RuGPT3\\ Medium\end{tabular}      & 0.468 & 0.01    & 0.372 / 0.461 & 0.598 & 0.706 / 0.308 & 0.505 & 0.642 & 0.669 & 0.634   & 0.23 / 0.224 \\ \hline
10   & \begin{tabular}[c]{@{}l@{}}RuGPT3\\ Small\end{tabular}       & 0.438 & -0.013  & 0.356 / 0.473 & 0.562 & 0.653 / 0.221 & 0.488 & 0.57  & 0.669 & 0.61    & 0.21 / 0.204 \\ \hline
11   & \begin{tabular}[c]{@{}l@{}}Baseline\\ TF-IDF1.1\end{tabular} & 0.434 & 0.06    & 0.301 / 0.441 & 0.486 & 0.587 / 0.242 & 0.471 & 0.57  & 0.662 & 0.621   & 0.26 / 0.252 \\ \hline
12   & \begin{tabular}[c]{@{}l@{}}Random\\ weighted\end{tabular}    & 0.385 & 0.0     & 0.319 / 0.374 & 0.48  & 0.45 / 0.071  & 0.483 & 0.528 & 0.597 & 0.52    & 0.25 / 0.247 \\ \hline
13   & \begin{tabular}[c]{@{}l@{}}majority\\ class\end{tabular}     & 0.374 & 0.0     & 0.217 / 0.484 & 0.498 & 0.0 / 0.0     & 0.513 & 0.587 & 0.669 & 0.503   & 0.25 / 0.247 \\ \hline
\end{tabular}
\end{sidewaystable}

\end{document}